\documentclass[11pt,a4paper]{article}
\usepackage[hyperref]{acl2021}
\usepackage{times}
\usepackage{latexsym}

\usepackage{graphicx}
\usepackage{booktabs}

\usepackage{microtype}

\aclfinalcopy %

\title{LAST at SemEval-2021 Task 1: Improving Multi-Word Complexity Prediction Using Bigram Association Measures}

\author{Yves Bestgen \\
  Laboratoire d'analyse statistique des textes - LAST \\
  Institut de recherche en sciences psychologiques \\
  Universit\'e catholique de Louvain \\
  Place Cardinal Mercier, 10 1348 Louvain-la-Neuve, Belgium \\
  \texttt{yves.bestgen@uclouvain.be} \\}

\date{}

\begin{document}
\maketitle
\begin{abstract}
This paper describes the system developed by the Laboratoire d'analyse statistique des textes (LAST) for the Lexical Complexity Prediction shared task at SemEval-2021. The proposed system is made up of a LightGBM model fed with features obtained from many word frequency lists, published lexical norms and psychometric data. For tackling the specificity of the multi-word task, it uses bigram association measures. Despite that the only contextual feature used was sentence length, the system achieved an honorable performance in the multi-word task, but poorer in the single word task. The bigram association measures were found useful, but to a limited extent.\end{abstract}

\section{Introduction}

For more than half a century, many studies have been carried out to collect norms about formal and semantic properties of words, such as frequency of use, spelling regularity, familiarity, age of acquisition, or emotional valence \citep{PRO99}. Some of these properties can be easily harvested through automatic counting procedures applied to corpora. Other properties, such as familiarity or emotional valence, are obtained by requiring participants, often more than ten, to rate the words on these dimensions. In psycholinguistics, these norms have been mainly used for selecting experimental materials \citep{WIL88}. In computational linguistics, they are used in opinion mining, in the evaluation of foreign language skills and in text simplification for instance \citep{PL08, KYL18}. Obtaining lexical norms that require human evaluations is extremely costly in time and resources, which greatly reduces their size. However, huge norms are essential in applications \citep{BES94}. This observation has led to the development of automatic techniques to extend such norms \citep{BES02, kamps-etal-2004-using, esuli-sebastiani-2006-sentiwordnet, BVI12}.
 
The Lexical Complexity Prediction (LCP) shared task at SemEval-2021 requires exactly the development of such techniques \citep{shardlow2021semeval}. It is indeed a question of estimating the lexical complexity, the degree of difficulty of the words in a text. This dimension is important in NLP applications for simplifying texts and assisting specific populations such as people with reading disabilities or who are learning a foreign language. A specificity of the LCP task is that it relates not only to words but also to multi-word expressions which are very rarely taken into account in norms and in automatic extension techniques \citep{BES14}. Another important feature of the task is that the target tokens were presented to human judges in context and that a significant number of them were presented in several different contexts. Human annotations are therefore likely to reflect the impact of the linguistic context on lexical complexity.
      
This paper describes the system proposed for this task by the Laboratoire d'analyse statistique des textes (LAST). It is based on a LightGBM model fed with features obtained from many word frequency lists, published lexical norms and psychometric data. For tackling the specificity of the multi-word task, it uses bigram association measures (such as Mutual Information) from research in lexicography \citep{church-hanks-1990-word} and in the automatic evaluation of texts written by English learners \citep{DUR09, BG14}. Despite that the only contextual feature used was sentence length, the system achieved an honorable performance in the multi-word task, ranking 9th out of 37 teams, but poorer in the single word task, ranking 26th out of 54 teams.

In the next section, the main characteristics of this challenge are summarized. The following section describes in detail the developed system. Finally, the results in the challenge are reported along with several analyzes performed to get a better idea of the factors that affect the system performance.

\section{Task and Materials}The organizers of the challenge have made available an updated version of the CompLex dataset \citep{shardlow2020complex} to the participating teams for developing their systems (i.e., the learning set). It consists of 8,083 single words and 1,616 bigrams, all of them presented in a one-sentence context. These sentences were taken from three English sources in almost equal proportion: biblical text, biomedical articles and proceedings of the European Parliament. The target words and bigrams were evaluated by several judges on a 5-point Likert scale depending on whether it seemed more or less easy to understand in this context. There were on average 25.75 annotations per instance \citep{shardlow2021predicting}. The complexity score for each target is the mean of these ratings. In this materials, a non-negligible proportion of the targets were presented several times in different sentences in order to assess the impact of this context on the complexity assessment. The test set, collected in the same way, consisted of 917 single words and 184 bigrams of which none of the targets were present in the learning set. The challenge measure was Pearson's linear correlation coefficient between human ratings and system predictions.

\section{System}
The first part of this section presents the features used to predict lexical complexity starting with those common to both tasks and ending with those specific to predicting the complexity of the multi-word expressions. Next, the procedure used to build the predictive models is described.

\subsection{Features}
\paragraph{Frequency Lists:}
I used the frequency of spelling forms calculated from corpora, but also a series of lists established by other researchers:
\begin{itemize}
\item The frequency in the Corpus of Contemporary American English (COCA), a balanced, 425-million word corpus of American English collected from 1990 to 2011 (\url{http://corpus.byu.edu/coca/}).
\item The frequency in the British National Corpus (BNC), a 100-million word collection of samples of written and spoken language designed to represent a wide cross-section of British English from the latter part of the 20th century (\url{http://www.natcorp.ox.ac.uk/corpus/}).
\item The Facebook frequency norms for American English and British English of \citet{HER17}, based on approximately 1 billion tokens for each English variety, obtained from publicly available English posts collected between November 2014 and January 2015.
\item The Rovereto Twitter Corpus frequency norms based on 75 millions tweets, for more than 1 billion tokens collected between December 2010 and July 2011 \citep{HER17}.
\item The USENET Orthographic Frequencies derived by \citet{SHA06} from a corpus of 7,781,959,860 words of USENET postings collected between October 2005 and August 2006.
\item The Hyperspace Analogue to Language (HAL) frequency norms provided by \citep{BAL07} for more that 40,000 words.
\item The frequency word list derived from the Google's ngram corpus available at \url{https://github.com/hackerb9/gwordlist}.
\end{itemize}
I also obtained the frequency of each target in each of the three corpora provided by the organizers as materials.

\paragraph{Lexical Norms and Psychometric Data:}
Lexical norms were mainly taken from the Glasgow Norms \citep{SCO19}. They contain the evaluation by human raters of 5,553 English words on the psycholinguistic dimensions of age of acquisition, arousal, concreteness, dominance, familiarity, gender association, imageability, semantic size and valence. I also used SemD, a measure of the semantic ambiguity of a word based on variability in its contextual usage \citep{HOF13}.
The psychometric data were taken from the English Lexicon Project \citep{BAL07}, a database that contains, for more than 40,000 words, the reaction time and average accuracy during lexical decision and naming tasks performed by many participants.

\paragraph{Other Features:}
Three binary features were used to encode the corpus from which the sentence is extracted, the initial analyzes having shown that it was more efficient than building three models, one per corpus. The only contextual feature taken into account was the sentence length in tokens.

\paragraph{Bigram Association Measures:}
These features, used only for the multi-word task, inform about the degree of association between the two target words according to a series of indices calculated on the basis of the frequency in a reference corpus of the bigram and that of the two words that compose it: pointwise mutual information and t-score \citep{church-hanks-1990-word}, z-score \citep{BER73}, log-likelihood Chi-square test \citep{dunning-1993-accurate}, simple-ll \citep{EVE09}, Dice coefficient \citep{KIL14} and the two delta-p \citep{KYL18}. \citet{BG14} refer to these features as \textit{collgrams} because they combine the strengths of both collocations (by using association scores) and n-grams (by using contiguous pairs of words). The justification for their use in the LCP task is given by works in foreign language learning which has shown that these indices can be used to assess the lexical richness of multi-word expressions present in texts written by English learners \citep{BG14, somasundaran-etal-2015-automated, BES17, BE19rfla}.

\subsection{Supervised Learning Software}
The regression models were built by the LightGBM open software \citep{LightGBM}, a well-known implementation of the gradient boosting decision tree approach. Compared to the multiple linear regression used  for this task by \citet{shardlow2020complex}, this type of model has the advantages of not requiring any feature preprocessing, such as a logarithmic transformation, since it is insensitive to monotonic transformations. It also allows a very effective overfit control thanks to its many parameters.

\subsection{Procedure}
The sentences were first lemmatized by the TreeTagger \citep{schmid1994treetagger}. The scores on the different lexical lists were attributed to the targets by a two-step procedure: on the basis of the orthographic form if it is found in the list or by using the lemma. The handling of missing values, which occurs when a word is not in a frequency list for example, has been left to the LightGBM default procedure. A large number of multi-word targets were given two values for many features by this procedure, one for each word. The corresponding features were doubled: the first encoding the minimum value and the second the maximum value.

The features used in the final models as well as LightGBM parameters were optimized by a 9-fold cross validation procedure. This led to the selection of the following features:
\begin{itemize}
\item For task 1, the length of the sentence and 12 features from the frequency lists, 10 from the lexical norms, and 8 from the psychometric data (i.e., average response latencies (raw and standardized), standard deviations, and accuracies for the lexical decision and naming tasks).
\item For task 2, the same features as in task 1 plus 3 features for the corpus of origin and 8 from the bigram association measures.
\end{itemize}

The same LightGBM parameters were used for both tasks. They were left at their default values except the followings: \textit{num\_iterations: 4800, max\_bin: 64, min\_data\_in\_bin: 10, lambda\_l2: 0.0175, bagging\_freq: 5, bagging\_fraction: 0.66, feature\_fraction: 0.09, learning\_rate: 0.0035, max\_depth: 7, min\_data\_in\_leaf: 7, num\_leaves: 11}. 

\section{Analyzes and Results}

\begin{table}
\begin{center}
\begin{tabular}{lrr}
\toprule
       &     \multicolumn{1}{c}{Test}  &   \multicolumn{1}{c}{CV}  \\ \midrule   
    &     \multicolumn{1}{c}{r}   &   \multicolumn{1}{c}{r} \\
Full System  &  0.753 &  0.810  \\ \midrule
        &     \multicolumn{1}{c}{Diff.}  &   \multicolumn{1}{c}{Diff.}  \\  
Length     & -0.002 &  -0.001   \\       
Frequencies & -0.013 &  -0.012   \\ 
Normes   & -0.022 &  -0.027   \\ \bottomrule
\end{tabular}
\caption{Difference in Pearson's r from the full system for the single word task when a set of features is removed (ablation approach).}
\end{center}
\end{table}

\begin{table*}
\centering
\begin{tabular}{cccccccrr}
\toprule
     &        &        &       &       &     &     \multicolumn{1}{c}{Test}   &  \multicolumn{1}{c}{CV}    \\ \midrule
Line Id & Sent. Length & Corpus Id & Norms & Freq. & Bigram  & \multicolumn{1}{c}{r} & \multicolumn{1}{c}{r} \\ \midrule
1 & x  &   x  &  x  &  x  & x &  0.842 &  0.799 \\ \midrule
    &    &      &     &     &   & Diff.  &  Diff.  \\                              
2 &    &   x  &  x  &  x  & x & -0.002 & -0.001 \\
3 & x  &      &  x  &  x  & x & -0.011 & -0.003 \\
4 & x  &   x  &     &  x  & x & -0.031 & -0.015 \\
5 & x  &   x  &  x  &     & x & -0.012 & -0.004 \\
6 & x  &   x  &  x  &  x  &   & -0.014 & -0.014 \\
7 &    &      &  x  &     &   & -0.096 & -0.055 \\
8 &    &      &     &  x  &   & -0.066 & -0.054 \\
9 &    &      &     &     & x & -0.176 & -0.161 \\ \bottomrule
\end{tabular}
\caption{Difference in Pearson's r from the full system for the multi-word task using the ablation approach.}
\end{table*}
\subsection{System Performance}
The system built to predict the lexical complexity of single words scored 0.7534 on the test material, ranking it 26th out of 54 teams, down 0.0352 from the best team. In the multi-word subtask, the system finished 9th out of 37 teams with a score of 0.8417. The best team got 0.8612.

The comparison of the results obtained on the test sets with those obtained by cross validation shows an unexpected difference between the two tasks. In the single word task, the correlation on the test set was lower by 0.053 compared to that obtained in CV (0.8064) while in the multi-word task this same correlation is higher by 0.042 compared to that obtained in CV (0.7996). It is also observed that the best systems which participated in the two tasks had superior performance on the multi-word task. If the difference in performance between the test sets and the CVs is not specific to the present system, this would suggest that the performance achieved in the multi-word task is rather overestimated, the test set being for some unknown reason relatively easy to predict. Although this is only a hypothesis which requires additional analyzes, it leads to not considering the multi-word task as being almost solved.

\subsection{Usefulness of the Different Types of Features}
In this section, the impact of the different types of features on the system performance is assessed using an ablation procedure. As the previous section indicated important differences between performance on the test set and by the CV approach, results are presented for these two evaluation procedures.

\paragraph{Single Word Task:}

Table 1 shows that the sentence length, the only contextual feature, is of little use. Norms and psychometric data are more useful than frequencies in corpora, but above all, these two sets of features provide very similar information since the removal of one as well as the other harms very little the model performance. These conclusions apply equally to the test set as to the CV.

\paragraph{Multi-Word Task:}

The system for multi-word expressions is based on five sets of features whose roles in its effectiveness are shown in Table 2. The absence of an "x" in a column indicates that this set of features has not been used in this version of the model. The first line of the table gives the performance of the system submitted for the challenge.

The length of the sentences [2] is much less useful than the features which identify the  corpus [3]. The comparison of the usefulness of the psychometric norms and data and the frequencies in corpora shows a contrast. When these sets are in turn excluded from the system, psychometric norms and data [4] are more useful than frequencies in corpora [5]. On the other hand, when used alone, frequencies [8] are more effective than psychometric norms and data [7]. It will be concluded that a greater part of the contribution of the frequency data is shared with other indices, most probably the bigram association measures.

The specific contribution of the bigram association measures [6] to the performance of the system is slightly greater than that of the frequencies in corpora. These features provide a gain of 0.014. Without it, the system would have been ranked 15th instead of 9th in this task. When used alone, however, bigram association measures [9] are much less effective than norms or frequencies.

\begin{table}
\begin{center}
\begin{tabular}{lrrr}
\toprule
Frequency& \multicolumn{2}{c}{Task} \\
of the target  & Single & Multi \\ \midrule
1       &      49.2 & 87.4 \\
2       &      19.9 &  7.8 \\
3       &      10.2 &  2.2 \\
4       &       6.2 &  0.8 \\
5       &      13.5 &  1.8 \\
6 or +  &       4.0 &  0.0 \\
Total (\%)  &    100.0 & 100.0 \\
Total nbr. &   3,850 & 1,479 \\  \bottomrule
\end{tabular}
\caption{Percentage of the target frequency in the two tasks.}
\end{center}
\end{table}

The effects of the different types of features are almost always more important when estimated on the test set rather than by CV. This could result from the initial difference in effectiveness between the two approaches. However, this phenomenon was not observed in the single-word task in which a difference in effectiveness was also observed. It is especially noted that the norms seem much more useful for the test set than for the CV. 

\paragraph{Potential Importance of the Context:}
The results presented above indicate that, in both tasks, sentence length is of little use. Taking better account of the context is undoubtedly a way to improve the system. This hypothesis is all the more likely as the role of context could explain the difference in performance between the two tasks of this system, but also of those of the other teams. Two observations support this hypothesis. Firstly, an analysis of the target frequencies in the two tasks, presented in Table 3, shows that there are much more repeated targets in the single-word task than in the multi-word task, a statistically significant difference for a Chi-square test ($p < 0.0001$).

\begin{figure}
 \begin{center}
  \includegraphics[width=7.7cm]{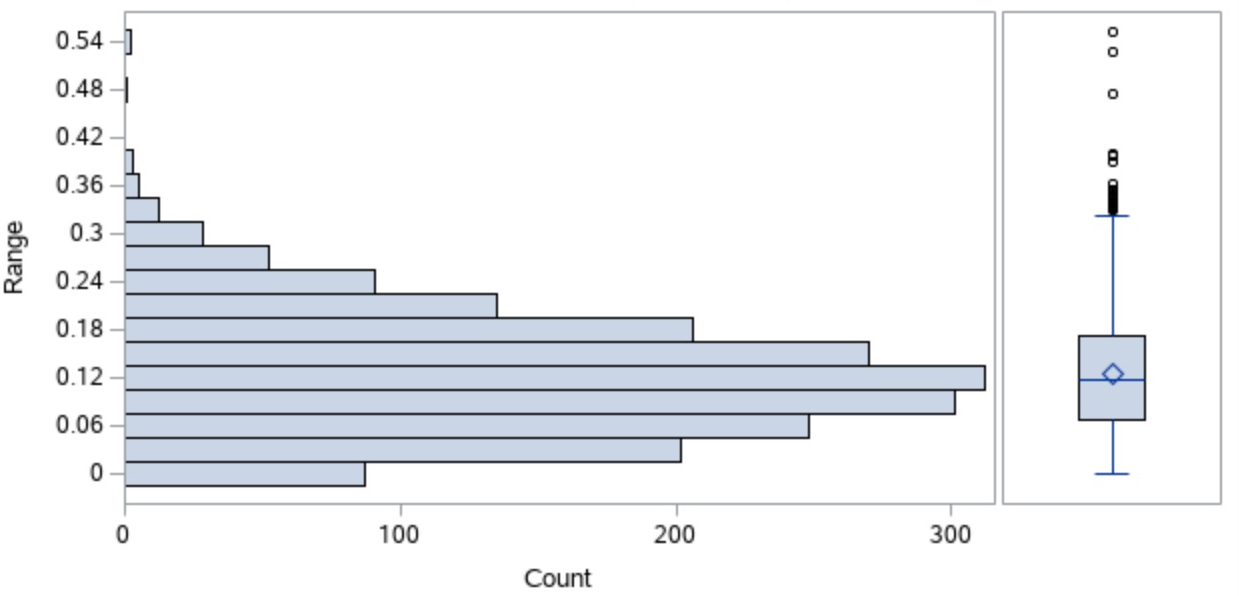}
 \end{center}
 \caption{Distribution of the range for the repeated targets in the single-word task.}
\end{figure}

Second, there are important differences between the human evaluations for the same target shown in different contexts. Figure 1 displays the distribution of the range (the difference between the maximum and the minimum values) of the complexity score for the repeated targets in the single-word task. The mean range is 0.125 and that 10\% of the repeated targets have a range greater than 0.224. Being able to take these differences into account in the single-word task could significantly improve the system, provided that the differences in evaluation for the same target are not just noise. Only an analysis of the inter-rater reliability for the repeated targets would make it possible to choose between these two options.

\section{Conclusion}
The models proposed for the LCP task were built by the LightGBM software mainly fed with norms and frequency features. It obtained an acceptable performance on the test set in the multi-word task on the basis of little contextual information, but less so in the single word task. The analyzes carried out by a CV approach showed, on the other hand, that the system is no better in the multi-word task. It is therefore possible or even probable that the better performance results from an overestimation of its effectiveness. The bigram association measures (aka CollGrams) have proven to be useful, but to a limited extent. 

Taking the context into account would probably have improved the system, especially for the single word task in which more than half of the targets were repeated. This hypothesis, however, is based on the assumption that differences between human ratings for the same target in different contexts are as reliable as their ratings for different targets. More generally, it would be interesting to explain the origin of the very important difference in performance between the two tasks, but that does not seem possible on the basis of the data I have access to.

\section*{Acknowledgments}
The author is a Research Associate of the Fonds de la Recherche Scientifique (FRS-FNRS).

\bibliographystyle{acl_natbib}
\bibliography{SemEval-2021Task1}

\begin{thebibliography}{30}
\expandafter\ifx\csname natexlab\endcsname\relax\def\natexlab#1{#1}\fi

\bibitem[{Balota et~al.(2007)Balota, Yap, Hutchison, Cortese, Kessler, Loftis,
  Neely, Nelson, Simpson, and Treiman}]{BAL07}
David~A. Balota, Melvin~J. Yap, Keith~A. Hutchison, Michael~J. Cortese, Brett
  Kessler, Bjorn Loftis, James~H. Neely, Douglas~L. Nelson, Greg~B. Simpson,
  and Rebecca Treiman. 2007.
\newblock \href {https://doi.org/10.3758/ BF03193014} {The {English} lexicon
  project}.
\newblock \emph{Behavior Research Methods}, 39:445--459.

\bibitem[{Berry-Rogghe(1973)}]{BER73}
Godelieve L.~M. Berry-Rogghe. 1973.
\newblock The computation of collocations and their relevance in lexical
  studies.
\newblock In Adam~J Aitken, Richard~W. Bailey, and Neil Hamilton-Smith,
  editors, \emph{The Computer and Literary Studies}. Edinburgh University
  Press.

\bibitem[{Bestgen(1994)}]{BES94}
Yves Bestgen. 1994.
\newblock Can emotional valence in stories be determined from words?
\newblock \emph{Cognition and Emotion}, 7:21--36.

\bibitem[{Bestgen(2002)}]{BES02}
Yves Bestgen. 2002.
\newblock D\'etermination de la valence affective de termes dans de grands
  corpus de textes.
\newblock In \emph{Actes du Colloque International sur la Fouille de Texte
  CIFT'02}, pages 81--94, Nancy. INRIA.

\bibitem[{Bestgen(2014)}]{BES14}
Yves Bestgen. 2014.
\newblock Construction automatique d'un lexique de n-grammes pour la fouille
  d'opinion : Faisabilit\'e et utilit\'e.
\newblock \emph{Document Num\'erique}, 17:103--123.

\bibitem[{Bestgen(2018)}]{BES17}
Yves Bestgen. 2018.
\newblock Beyond single-word measures: {L2} writing assessment, lexical
  richness and formulaic competence.
\newblock \emph{System}, 69:65--78.

\bibitem[{Bestgen(2019)}]{BE19rfla}
Yves Bestgen. 2019.
\newblock Evaluation de textes en anglais langue ètrangère et séries
  phraséologiques : comparaison de deux procédures automatiques librement
  accessibles.
\newblock \emph{Revue française de linguistique appliquée}, 24:81--94.

\bibitem[{Bestgen and Granger(2014)}]{BG14}
Yves Bestgen and Sylviane Granger. 2014.
\newblock Quantifying the development of phraseological competence in {L2
  English} writing: An automated approach.
\newblock \emph{Journal of Second Language Writing}, 26:28--41.

\bibitem[{Bestgen and Vincze(2012)}]{BVI12}
Yves Bestgen and Nadja Vincze. 2012.
\newblock \href {https://doi.org/10.3758/s13428-017-0924-4} {Checking and
  bootstrapping lexical norms by means of word similarity indexes}.
\newblock \emph{Behavior Research Methods}, 44:998--1006.

\bibitem[{Church and Hanks(1990)}]{church-hanks-1990-word}
Kenneth~Ward Church and Patrick Hanks. 1990.
\newblock \href {https://www.aclweb.org/anthology/J90-1003} {Word association
  norms, mutual information, and lexicography}.
\newblock \emph{Computational Linguistics}, 16(1):22--29.

\bibitem[{Dunning(1993)}]{dunning-1993-accurate}
Ted Dunning. 1993.
\newblock \href {https://www.aclweb.org/anthology/J93-1003} {Accurate methods
  for the statistics of surprise and coincidence}.
\newblock \emph{Computational Linguistics}, 19(1):61--74.

\bibitem[{Durrant and Schmitt(2009)}]{DUR09}
Philip Durrant and Norbert Schmitt. 2009.
\newblock To what extent do native and non-native writers make use of
  collocations?
\newblock \emph{International Review of Applied Linguistics in Language
  Teaching}, 47:157--177.

\bibitem[{Esuli and Sebastiani(2006)}]{esuli-sebastiani-2006-sentiwordnet}
Andrea Esuli and Fabrizio Sebastiani. 2006.
\newblock \href {http://www.lrec-conf.org/proceedings/lrec2006/pdf/384_pdf.pdf}
  {{SENTIWORDNET}: A publicly available lexical resource for opinion mining}.
\newblock In \emph{Proceedings of the Fifth International Conference on
  Language Resources and Evaluation ({LREC}{'}06)}, Genoa, Italy. European
  Language Resources Association (ELRA).

\bibitem[{Evert(2009)}]{EVE09}
Stefan Evert. 2009.
\newblock Corpora and collocations.
\newblock In Anke L{\"u}deling and Merja Kyt{\"o}, editors, \emph{Corpus
  Linguistics. An International Handbook}, pages 1211--1248. Mouton de Gruyter.

\bibitem[{Herdagdelen and Marelli(2017)}]{HER17}
Amac Herdagdelen and Marco Marelli. 2017.
\newblock \href {https://doi.org/10.1111/cogs.12392} {Social media and language
  processing: How facebook and twitter provide the best frequency estimates for
  studying word recognition}.
\newblock \emph{Cognitive Science}, 41:976--995.

\bibitem[{Hoffman et~al.(2013)Hoffman, Ralph, and Rogers}]{HOF13}
Paul Hoffman, Matthew A.~Lambon Ralph, and Timothy~T. Rogers. 2013.
\newblock \href {https://doi.org/10.3758/s13428-012-0278-x} {Semantic
  diversity: A measure of semantic ambiguity based on variability in the
  contextual usage of words}.
\newblock \emph{Behavior Research Methods}, 45:998--1006.

\bibitem[{Kamps et~al.(2004)Kamps, Marx, Mokken, and
  de~Rijke}]{kamps-etal-2004-using}
Jaap Kamps, Maarten Marx, Robert~J. Mokken, and Maarten de~Rijke. 2004.
\newblock \href {http://www.lrec-conf.org/proceedings/lrec2004/pdf/734.pdf}
  {Using {W}ord{N}et to measure semantic orientations of adjectives}.
\newblock In \emph{Proceedings of the Fourth International Conference on
  Language Resources and Evaluation ({LREC}{'}04)}, Lisbon, Portugal. European
  Language Resources Association (ELRA).

\bibitem[{Ke et~al.(2017)Ke, Meng, Finley, Wang, Chen, Ma, Ye, and
  Liu}]{LightGBM}
Guolin Ke, Qi~Meng, Thomas Finley, Taifeng Wang, Wei Chen, Weidong Ma, Qiwei
  Ye, and Tie-Yan Liu. 2017.
\newblock \href
  {http://papers.nips.cc/paper/6907-lightgbm-a-highly-efficient-gradient-boosting-decision-tree.pdf}
  {{LightGBM}: A highly efficient gradient boosting decision tree}.
\newblock In I.~Guyon, U.~V. Luxburg, S.~Bengio, H.~Wallach, R.~Fergus,
  S.~Vishwanathan, and R.~Garnett, editors, \emph{Advances in Neural
  Information Processing Systems 30}, pages 3146--3154. Curran Associates, Inc.

\bibitem[{Kilgarriff et~al.(2014)Kilgarriff, Baisa, Bu{\v s}ta, Jakub{\'\i}{\v
  c}ek, Kov{\'a}{\v r}, Michelfeit, Rychl{\'y}, and Suchomel}]{KIL14}
Adam Kilgarriff, V{\'\i}t Baisa, Jan Bu{\v s}ta, Milo{\v s} Jakub{\'\i}{\v
  c}ek, Vojt{\v e}ch Kov{\'a}{\v r}, Jan Michelfeit, Pavel Rychl{\'y}, and
  V{\'\i}t Suchomel. 2014.
\newblock The {Sketch Engine}: ten years on.
\newblock \emph{Lexicography}, 1(1):7--36.

\bibitem[{Kyle et~al.(2018)Kyle, Crossley, and Berger}]{KYL18}
Kristopher Kyle, Scott Crossley, and Cynthia Berger. 2018.
\newblock \href {https://doi.org/10.3758/s13428-017-0924-4} {The tool for the
  automatic analysis of lexical sophistication ({TAALES}): version 2.0}.
\newblock \emph{Behavior Research Methods}, 50:1030--1046.

\bibitem[{Pang and Lee(2008)}]{PL08}
Bo~Pang and Lillian Lee. 2008.
\newblock \href {https://doi.org/10.1007/BF02259727} {Opinion mining and
  sentiment analysis}.
\newblock \emph{Foundations and Trends in Information Retrieval}, 2(1):1--135.

\bibitem[{Proctor and Vu(1999)}]{PRO99}
Robert~W. Proctor and Kim-Phuong~L. Vu. 1999.
\newblock \href {https://doi.org/10.3758/s13428-018-1099-3} {Index of norms and
  ratings published in the psychonomic society journals}.
\newblock \emph{Behavior Research Methods}, 31:659--667.

\bibitem[{Schmid(1994)}]{schmid1994treetagger}
Helmutt Schmid. 1994.
\newblock {Probabilistic part-of-speech tagging using decision trees}.
\newblock In \emph{International Conference on New Methods in Language
  Processing}, pages 44--49.

\bibitem[{Scott et~al.(2019)Scott, Keitel, Becirspahic, Yao, and
  Sereno}]{SCO19}
Graham~G. Scott, Anne Keitel, Marc Becirspahic, Bo~Yao, and Sara~C. Sereno.
  2019.
\newblock \href {https://doi.org/10.3758/s13428-018-1099-3} {The {Glasgow}
  norms: Ratings of 5,500 words on nine scales}.
\newblock \emph{Behavior Research Methods}, 51:1258--1270.

\bibitem[{Shaoul and Westbury(2006)}]{SHA06}
Cyrus Shaoul and Chris Westbury. 2006.
\newblock {USENET} orthographic frequencies for 111,627 {E}nglish words.

\bibitem[{Shardlow et~al.(2020)Shardlow, Cooper, and
  Zampieri}]{shardlow2020complex}
Matthew Shardlow, Michael Cooper, and Marcos Zampieri. 2020.
\newblock Complex: A new corpus for lexical complexity predicition from likert
  scale data.
\newblock In \emph{Proceedings of the 1st Workshop on Tools and Resources to
  Empower People with REAding DIfficulties (READI)}.

\bibitem[{Shardlow et~al.(2021{\natexlab{a}})Shardlow, Evans, Paetzold, and
  Zampieri}]{shardlow2021semeval}
Matthew Shardlow, Richard Evans, Gustavo Paetzold, and Marcos Zampieri.
  2021{\natexlab{a}}.
\newblock Semeval-2021 task 1: Lexical complexity prediction.
\newblock In \emph{Proceedings of the 14th International Workshop on Semantic
  Evaluation (SemEval-2021)}.

\bibitem[{Shardlow et~al.(2021{\natexlab{b}})Shardlow, Evans, and
  Zampieri}]{shardlow2021predicting}
Matthew Shardlow, Richard Evans, and Marcos Zampieri. 2021{\natexlab{b}}.
\newblock \href {http://arxiv.org/abs/2102.08773} {Predicting lexical
  complexity in english texts}.

\bibitem[{Somasundaran et~al.(2015)Somasundaran, Lee, Chodorow, and
  Wang}]{somasundaran-etal-2015-automated}
Swapna Somasundaran, Chong~Min Lee, Martin Chodorow, and Xinhao Wang. 2015.
\newblock \href {https://doi.org/10.3115/v1/W15-0605} {Automated scoring of
  picture-based story narration}.
\newblock In \emph{Proceedings of the Tenth Workshop on Innovative Use of {NLP}
  for Building Educational Applications}, pages 42--48, Denver, Colorado.
  Association for Computational Linguistics.

\bibitem[{Wilson(1988)}]{WIL88}
Michael Wilson. 1988.
\newblock \href {https://doi.org/10.3758/BF03202594} {{MRC} psycholinguistic
  database: Machine-usable dictionary, version 2.00}.
\newblock \emph{Behavior Research Methods}, 20:6--10.

\end{thebibliography}

\end{document}